\documentclass[tinyml]{acmart}

\AtBeginDocument{%
  \providecommand\BibTeX{{%
    \normalfont B\kern-0.5em{\scshape i\kern-0.25em b}\kern-0.8em\TeX}}}

\setcopyright{rightsretained}
\copyrightyear{2023}
\acmYear{2023}

\usepackage{booktabs}
\usepackage{tikz}
\usepackage{pgfplots}
\usepackage{adjustbox}

\begin{document}

\title{SSS3D: Fast Neural Architecture Search For
Efficient Three-Dimensional Semantic
Segmentation}






\author{Olivier Therrien, Marihan Amein, Zhuoran Xiong, Warren J. Gross, Brett H. Meyer}
\affiliation{%
  \institution{McGill University}
  \city{Montr\'eal}
  \state{Qu\'ebec}
  \country{Canada}
}
\email{{olivier.therrien, marihan.amein, zhuoran.xiong}@mail.mcgill.ca}
\email{{warren.gross, brett.meyer}@mcgill.ca}


\begin{abstract}
We present SSS3D, a fast multi-objective NAS framework designed to find computationally efficient 3D semantic scene segmentation networks.
It uses RandLA-Net, an off-the-shelf point-based network, as a super-network to enable weight sharing and reduce search time by 99.67\% for single-stage searches.
SSS3D has a complex search space composed of sampling and architectural parameters that can form $2.88\times10^{17}$ possible networks.
To further reduce search time, SSS3D splits the complete search space and introduces a two-stage search that finds optimal subnetworks in 54\% of the time required by single-stage searches.
\end{abstract}


\maketitle

\section{Introduction}

3D point clouds are increasingly used in autonomous driving and in other areas like robotics, industrial control, augmented reality, and medical image analysis~\cite{he2021deep}.
Light Detection and Ranging (LiDAR) sensors, for instance, are often used in autonomous cars~\cite{hecht2018lidar} for their ability to produce precise distance measurements, and for being less affected by the lighting, visibility, depth, and FOV of the scene than 2D images~\cite{cheng20212}.
These sensors generate point cloud representations of their surrounding environment, that are rich in details but are difficult to interpret given their unordered structure.

Just like in 2D, the state-of-the-art (SOTA) architectures for 3D semantic segmentation~\cite{zhu2021cylindrical, qian2022pointnext} tend to be complex, and are generally optimized for accuracy.
They also often make use of more complex operations like 3D and sparse convolutions.
Their size limits their performance when deployed on embedded platforms and the operations they use can also limit the number of embedded platforms they can be deployed on.
Because of the nature of the tasks requiring 3D semantic scene segmentation, competitive computationally efficient architectures have been developed, e.g., 2DPASS~\cite{yan20222dpass}, JSRC-Net~\cite{yan2021sparse} and SPVNAS~\cite{tang2020searching}.
However, these still use operations that are not supported on some platforms. 

We propose Supernet Semantic Segmentation 3D (SSS3D), a fast multi-objective NAS framework that uses the elitist multi-objective genetic algorithm (GA) NSGA-II~\cite{deb2002fast} to optimize RandLA-Net \cite{hu2020randla} for accuracy, size and computational cost.
RandLA-Net is a point-based 3D semantic segmentation network with an encoder-decoder structure.
SSS3D searches for sets of architectural and sampling parameters that reduce computational complexity without degrading accuracy. 
Architectural hyperparameters modify the internal structure of the supernet, while sampling parameters do not.
Such configurations could also benefit SOTA networks like~\cite{choe2021pointmixer, qian2022pointnext} that are similar to RandLA-Net.

SSS3D employs weight sharing inspired by Once-for-All~\cite{cai2019once}, reducing overall training time by 99.67\% by eliminating candidate training.
An early stopping criterion is used to reduce the evaluation time of the sampled architectures during the search.
At the end of the search, fine-tuning is used on Pareto dominant architectures to increase their performance.
We observe that using two-stage search, optimizing sampling parameters first, and architecture parameters second, reduces search time by 46\% compared to single-stage search and minimizes multiple computational cost metrics.

SSS3D finds efficient variations of RandLA-Net in 1.04 to 2.69 GPU days.
On S3DIS~\cite{S3DIS}, RandLA-Net has 5M parameters and uses 17G FLOPs to achieve 62.78\% mIoU.
Model SAP-1 increases accuracy by 0.27\% after fine-tuning but uses only 53.81\% and 88.43\% of RandLA-Net’s parameters and FLOPs respectively.
SF-1 has a mIoU of 62.91\% but reduces FLOPs by 11.16\% compared to RandLA-Net.
SAP-2 has the lowest parameter count of models found, with 48.34\% of the original parameters, and achieves 61.58\% mIoU after fine-tuning.
AFP-3 has the least FLOPs, with a 62.36\% reduction, and achieves an accuracy of 59.28\% after fine-tuning.

\section{Related Work}

SSS3D is a fast multi-objective NAS framework that uses NSGA-II~\cite{deb2002fast}, an elitist genetic algorithm (GA), to optimize the inner structure and the sampling choices of its RandLA-Net super-network.
Our goals are to reduce the search time of NAS and to find computationally efficient networks that use basic operations making them deployable on most embedded platforms.

\subsection{3D Semantic Scene Segmentation}

Traditional 2D semantic image segmentation architectures~\cite{chen2018encoder, zhao2017pyramid} are not as successful on the 3D version of the task because their convolutions exploit local spatial relations between neighboring pixels.
They can not be applied directly to unordered and sometimes sparse structure like 3D point clouds without degrading accuracy~\cite{li2021multi}.
3D representation-based semantic segmentation networks~\cite{choy20194d, cheng20212, zhu2021cylindrical, yan20222dpass, tang2020searching} transform the point cloud input into a 3D structure.
As such, they use complex operations like sparse, 3D, or higher dimension convolutions to extract features, limiting their deployment to platforms that support them.
Projection-based networks~\cite{boulch2018snapnet, wu2018squeezeseg, wu2019squeezesegv2, xu2020squeezesegv3} solve this problem by projecting the point clouds into 2D space to extract and interpret features.
However, projection loses information, such as the geometric relations between points, degrading accuracy~\cite{zhu2021cylindrical}.
Point-based networks like~\cite{yan2021sparse, choe2021pointmixer, zhao2021point, qian2022pointnext} use the complete unordered point cloud~\cite{he2021deep}, preventing information loss during preprocessing.
The downside is that they require many random memory accesses because of the unordered nature of the point cloud~\cite{tang2020searching}.
RandLA-Net~\cite{hu2020randla} addresses this issue by using random sampling and an efficient context aggregation module.

\subsection{Neural Architecture Search}

NAS frameworks for 2D computer vision tasks~\cite{cai2019once, guo2020single, stamoulis2019single, shaw2019squeezenas, chen2018searching} are not well suited for 3D versions of the same task.
The problem changes with the addition of the third dimension, and different kinds of parameters need to be explored.
NAS for 3D medical image segmentation~\cite{8885744, bae2019resource, wong2019segnas3d} are not directly applicable to point clouds of 3D scenes either, because the format of 3D medical data is similar to 2D images~\cite{tang2020searching}.
3D scene point cloud problems are also larger than 3D objects point clouds~\cite{tang2020searching}.

Very few hardware-aware NAS frameworks for 3D point cloud semantic segmentation exist.
LC-NAS~\cite{li2020lc} is a gradient-based framework that uses a latency regressor to predict the latency of the searched network.
Its performance is tested for classification and object part segmentation only.
Gradient-based frameworks are limited by the feasibility of presenting the search space and the different objectives, like computational cost metrics, in a differentiable form~\cite{amein2022multi}.
GA-based frameworks are generally faster than reinforcement learning based frameworks while making it easier to design the search space than gradient-based frameworks~\cite{amein2022multi}.
3D-NAS \cite{tang2020searching} is a GA-based framework that searches only the architectural parameters of its supernet consisting of 3D convolutions.
It is a constraint-based framework that simply drops potential networks that do not meet the hardware constraints.
SSS3D uses a multi-objective GA instead to find Pareto-optimal networks that can satisfy different levels of hardware constraints and searches the sampling and architectural parameters of its supernet made of 2D convolutions.

To date, the two-stage search approach is used in 2D NAS frameworks only.
Frameworks like~\cite{hu2020count}~and~\cite{liu2019auto} create a two-stage differentiable search space that optimizes micro and macro-level parameters simultaneously.
In HURRICANE~\cite{zhang2020fast}, the first stage searches for efficient operations in half of the network and the second stage searches the other half.
The two stages happen at the same time.
BP-NAS~\cite{liu2020block} splits the search like SSS3D.
It searches for the structure of the network’s blocks in the first stage and the connections between the blocks in the second.
SSS3D searches the sampling parameters in the first stage and the architectural parameters in the second stage using a fixed supernet structure to facilitate weight sharing and reduce search time.

\section{Supernet Semantic Segmentation 3D}

\begin{figure*}[t]
  \centering
  \includegraphics[width=0.9\textwidth]{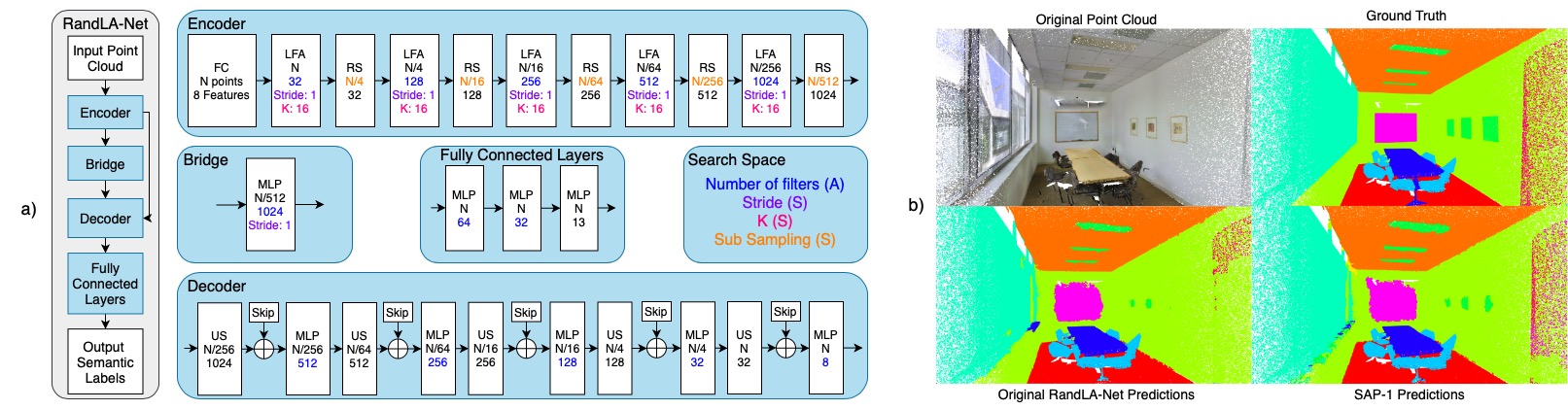}
  \caption{a) Structure of RandLA-Net~\cite{hu2020randla} and the sampling (S) and architectural (A) parameters of SSS3D's search space. In a block, the first line indicates the operation done, the second is the number of points used, the third is the number of output features and the other lines are extra parameters of that operation. FC: Fully Connected, LFA: Local Feature Aggregation, RS: Random Sampling, MLP: Multi-Layer Perceptron, US: Upsampling, Skip: Skip Connection. b) Predictions made by the original RandLA-Net and SAP-1 for a room in area five of S3DIS~\cite{S3DIS}. The original point cloud and the ground truth are also shown.}
  \label{searchSpace}
\end{figure*}

We present SSS3D: a fast multi-objective NAS framework that searches for computationally efficient 3D semantic scene segmentation networks using the elitist genetic algorithm NSGA-II~\cite{deb2002fast}.
To reduce the candidate evaluation time and total search time, SSS3D uses pre-trained weights from its RandLA-Net super-network, illustrated in Figure~\ref{searchSpace}(a), and early stopping.
SSS3D divides search into two stages to reduce search time even more.
The first stage searches the sampling parameters to find subnetworks that are fast, require fewer memory accesses, and maintain accuracy.
In the second stage, the sampling parameters are fixed to a combination found in the first stage, and the architectural parameters are searched to reduce model size.
The subnetworks on the final Pareto optimal front of the single and two-stage searches are fine-tuned.

The SSS3D search algorithm starts with a population of size $N$ that has $N-1$ randomly sampled architectures from the search space.
RandLA-Net itself is the last member of the population, to guide the search by providing a high-performance network early.
The search then runs for $G$ generations.
Half of each subsequent generation's population is selected from the current using NSGA-II’s criteria, which favours non-dominated candidates.
The second half of the population is composed of networks created using crossover and mutation of networks from the first half.

\subsection{Creating Subnetworks}
\subsubsection{RandLA-Net}

SSS3D uses RandLA-Net~\cite{hu2020randla} as a supernet to constrain the high-level network architecture of the candidates and to enable weight sharing to reduce training cost at evaluation time.
RandLA-Net is a point-based network for 3D semantic scene segmentation that has an encoder-decoder architecture.
It uses the raw point cloud as input and has a sampling strategy to increase its efficiency by progressively reducing the number of point features processed.
RandLA-Net uses random point sampling at every encoder block to reduce the size of the input point cloud.
To avoid the loss of important information, it uses a novel local feature aggregation module, LFA in Figure~\ref{searchSpace}a), before every sampling operation, increasing the receptive field of the points and preserving the geometric details of the pruned points~\cite{hu2020randla}.
The LFA module is mostly made of 2D 1x1 convolutions.
Upsampling in the decoder provides a class prediction for every point of the input cloud.
RandLA-Net achieves 70\% mIoU using 5M parameters and 17G FLOPs on the 6-fold cross validation of S3DIS~\cite{S3DIS}.
It also achieves 77.4\% mIoU and 53.9\% mIoU on Semantic3D \cite{hackel2017semantic3d} and SemanticKITTI \cite{behley2019semantickitti}.

SSS3D searches for more efficient implementations of RandLA-Net.
It is a lightweight architecture, but different sampling strategies can be explored to make it more efficient.
The size of the network can also be reduced to fit on even more constrained hardware platforms by optimizing the number of filters.
SSS3D could be applied to search other supernets, as many competitive architectures have similar structure.
RandLA-Net has the same encoder-decoder structure as point-based networks like~\cite{choe2021pointmixer} and~\cite{qian2022pointnext}.
It is also used as a backbone architecture in other works like~\cite{wang2022semantic} and~\cite{geng2021multi}.

\subsubsection{Search Space}

The search space of SSS3D is built around the structure of RandLA-Net and has two types of parameters.
\emph{Sampling parameters} are hyperparameters that control network operation without changing its structure and aim to reduce memory accesses and FLOPs.
\emph{Architectural parameters}, in contrast, alter the supernet’s inner structure and target a reduction in network size in memory.

\begin{table}[b]
\centering
\caption{The sampling (S) and architectural (A) parameters of SSS3D and their possible values.}
\small
\begin{tabular}{lr}
    \toprule
    Parameters & Values \\
    \midrule
    Filter Ratio (A) & 1.0, 0.8, 0.6, 0.4\\
    Stride (S) & 1, 2, 3, 4\\ 
    K (S) & 16, 18, 20, 22\\
    Subsampling (S) & 2, 4, 6, 8\\
    \bottomrule
\end{tabular}
\label{tableSP}
\end{table}

Because of the unordered nature of raw point clouds, it is crucial to limit the number of points used during inference for an efficient point-based 3D semantic segmentation network.
Standard point-based networks like PointNet~\cite{qi2017pointnet} spend up to 90\% of their runtime on structuring the irregular data instead of feature extraction~\cite{tang2020searching}.
Sampling parameters are a type of NAS hyperparameters unique to 3D point cloud tasks.
Heavy pruning of a 2D image, for example, could never produce results similar to the original network.
The stride of the convolutions in the encoder and the bridge, the number of neighbours used during LFA, and the subsampling ratios between every step of the encoder, are the sampling parameters explored by SSS3D (Figure~\ref{searchSpace}a) and Table~\ref{tableSP}).
At each encoder step, a bunch of points are fed to 1x1 convolutional filters producing a 1x1 output for every point and every convolutional filter.
Increasing convolutional stride reduces the number of point features that are fed to the 1x1 filters, reducing the number of computations, while the output dimension remains the same.
In 2D, the convolutional strides could not be used as a sampling parameter as it modifies the size of the output for convolutional filters greater than 1x1.
LFA uses the KNN algorithm to gather the point features from the neighbours of the points before sampling.
Higher $K$ implies information is collected from more points, reducing the risk of not sampling points that were excluded from the neighbourhood of other points hence increasing network's accuracy.
The subsampling ratio determines the ratio of points that are randomly sampled from one stage of the encoder to the next.
It has a direct effect on the number of computations required by the architecture, since fewer points means less computations.
The subsampling ratio and the $K$ value have an interesting dynamic:
the effect on accuracy of sampling more points can be diminished by larger $K$, and vice versa.

The architectural parameters we explore are the number of convolutional filters used in each part of RandLA-Net.
As in~\cite{cai2019once}, the L1 norm of the convolutional filters’ weights is used to determine the importance of the filters; the least important filters are pruned to match the candidate's targeted filter count.
Our hypothesis is that RandLA-Net, and other similar networks~\cite{qian2022pointnext,choe2021pointmixer}, are overparameterized.
Reducing the number of filters is a straightforward way to reduce complexity, often with minimal impact on accuracy. 
This is critically important when deploying in hardware-constrained environments, as pruning improves inference latency, and reduces memory footprint. 

\subsection{Subnetwork Evaluation}

Training is not required when evaluating subnetworks with differing sampling parameters: the number of weights are not changed, just which points are used for which computations.
Changes in architectural parameters, however, change the internal structure of the candidate relative to the supernet.
We use progressive shrinking (PS)~\cite{cai2019once} to train all the possible subnetworks.

PS starts by training the largest network in the search space, and then progressively fine-tunes it to support smaller and smaller subnetworks~\cite{cai2019once}.
In SSS3D, the initial training step is not required: pre-trained RandLA-Net is the biggest network possible. 
However, subsequent fine-tuning allows SSS3D to have a training cost of O(1), instead of O(n) (if every sampled subnetwork was trained from scratch).
During fine-tuning, different subnetworks are sampled for every batch of training data, trained, and then the supernet’s weights are updated using knowledge distillation~\cite{hinton2015distilling}.

SSS3D uses early stopping when computing accuracy, to reduce evaluation time.
The time required to evaluate the accuracy of candidates is a bottleneck in NAS. 
We observed that the sampled networks with poor accuracy at the start of the test set also had poor accuracy on the rest of the test set.
The average standard deviation of a network’s accuracy on the test set is 0.034, suggesting that accuracy is almost constant over the test set's batches.
To avoid wasting time evaluating a design that is uncompetitive, the accuracy of the networks are checked after 25\% of the test samples have been passed.
If the mean accuracy is less than 30\% at that point, network evaluation is stopped.
This can save, on average, 3.51 GPU minutes per candidate compared to a full evaluation.
The threshold is set at 30\% accuracy to eliminate networks with terrible performance only.
It is important to evaluate subnetworks that do not perform as well as the supernet, because it allows the search to learn about the effect of the value of its hyperparameters.

\section{Experimental Setup}

We conducted two experiments on S3DIS~\cite{S3DIS} to demonstrate the speed at which SSS3D can discover computationally efficient and hardware-aware 3D point-based semantic segmentation networks.
The first experiment searches the complete search space, both sampling and architectural parameters, in a single stage, to find subnetworks that minimize mIoU error and computational cost.
The second experiment tests a two-stage approach that searches sampling parameters first, and architectural parameters second.
These experiments were done using an Nvidia GeForce GTX 1080 Ti GPU.

\subsection{Evaluation Metrics}

The accuracy metric used in SSS3D is the mean intersection over union (mIoU) error.
MIoU is a standard metric used in semantic segmentation that measures the overlap between the predictions and the ground truth labels~\cite{behley2019semantickitti}.
MIoU is calculated by dividing the true positives by the sum of the true positives, false positives, and false negatives, for every class in the dataset, and then taking the mean of that value across all classes.
The mIoU error is obtained by subtracting the mIoU from one.

Two computational metrics are used in SSS3D: number of parameters and number of FLOPs.
The number of parameters is a good indicator of the size of the network in memory but also affects the number of computations done by the network, affecting its latency.
The number of floating point operations is also used to get an idea of a network’s latency.
The more operations it does, the higher the latency and vice versa.
Less operations can also mean less memory accesses.
Minimizing these metrics while maintaining competitive mIoU should result in more efficient variations of RandLA-Net.

\subsection{S3DIS Dataset}

The dataset used by SSS3D for training, fine-tuning and evaluation is the Stanford 3D indoor scene dataset (S3DIS)~\cite{S3DIS,S3DIS2}.
It is an indoor 3D point cloud dataset made of scans of large buildings with various architectural styles~\cite{S3DIS2}.
It has scans from a total of six areas containing 271 rooms in total.
The dataset has 13 semantic categories: ceiling, floor, wall, beam, column, window, door, chair, table, bookcase, sofa, board and clutter.
In~\cite{S3DIS}, the authors present different training and testing splits for the dataset.
In all of our experiments the first fold is used.
Training is done on areas one, two, three, four, and six, and testing is done on area five.

\subsection{Search Hyperparameters}

SSS3D searches for efficient RandLA-Net variations using sampling and architectural parameters, as shown in Table~\ref{tableSP}.
Stride, $K$, and subsampling, form the sampling search space.
These three parameters are applied to the five stages of the RandLA-Net encoder; each stage can use different values.
Stride is also applied to the 1x1 convolution inside the bridge.
The sampling search space has more than 4B possible designs. 

The architectural search space consists of changes in the number of filters (by specifying a ratio of the total available) in the network.
The filter ratio is set independently at 13 places in the RandLA-Net architecture, including the encoder, bridge, decoder, and first two fully-connected layers at the end, as shown in blue in Figure~\ref{searchSpace}a).
The ratio is multiplied by the original number of convolutional filters in the corresponding block; excess filters are pruned.
With four choices for each filter ratio, there are consequently more than 67M possible designs in the architectural search space. 
This makes the use of PS a key component of SSS3D, as only one tuning of RandLA-Net's weights is required for all possible candidate subnetworks. 
The complete search space is the combination of the sampling and architectural search space and has nearly $3\times10^{17}$ parameters. 

Single-stage searches use a population of 15 and search for 60 generations.
For two-stage searches, both stages use a population of 12; the first stage searches the sampling space for 20 generations, and the second stage searches the architectural space for 15.
These numbers were determined experimentally by observing a variation of the \emph{hyperarea difference} (HD)~\cite{wu2001metrics} metric.
It compares the area dominated by two consecutive Pareto optimal fronts (POF).
When the HD values get smaller and stabilize, the POF is no longer changing substantially and the search can be stopped.

\subsection{PS and Final Fine-Tuning}

The progressive shrinking algorithm has hyperparameters that dictate the growth of the pool from which subnetworks are subsampled during the tuning phase.
SSS3D uses a dynamic batch size of one~\cite{cai2019once}. 
The dynamic batch size corresponds to the number of subnetworks that are sampled from the search space for a single batch of the training data.
SSS3D uses a knowledge distillation ratio of one~\cite{cai2019once}.
SSS3D performs PS tuning in a single step.
When the architectural design space is bigger, a multi-step approach can be used by adding some of the possible values of a parameter in the search space at every step.
The Adam optimizer is used for training 60 epochs with a learning rate of 2.5$e^{-3}$, and a batch size of three.

Final fine-tuning is performed on the most interesting Pareto-optimal candidates: e.g., ones that minimize cost, and mIoU degradation relative to the supernet.
Final fine-tuning uses an Adam optimizer for ten epochs, a starting learning rate of 1$e^{-4}$, a learning rate decay of 0.95, and a batch size of three.

\section{Results}

\subsection{Single-stage Search}

\begin{figure}[!t]
\centering
    \begin{tikzpicture}[scale=0.9]
        \begin{axis}[
            enlargelimits=false,
            xlabel={Params (M)},
            ylabel={Mean IoU Error (\%)},
            tick align=outside, legend pos=north east,
            xmin= 2, xmax=5.5,
            ymin= 30, ymax=100,
        ]
        \addplot[
            color=black]
        table[x=Params, y=Error]
        {data/S3DIS/Params/gen1.tex};
        \addplot[
            color=black,
            only marks,
            mark=halfcircle,
            mark size=2pt]
        table[x=Params, y=Error]
        {data/S3DIS/Params/gen1.tex};
        \addplot[
            color=brown]
        table[x=Params, y=Error]
        {data/S3DIS/Params/gen10.tex};
        \addplot[
            color=brown,
            only marks,
            mark=halfcircle,
            mark size=2pt]
        table[x=Params, y=Error]
        {data/S3DIS/Params/gen10.tex};
        \addplot[
            color=purple]
        table[x=Params, y=Error]
        {data/S3DIS/Params/gen20.tex};
        \addplot[
            color=purple,
            only marks,
            mark=halfcircle,
            mark size=2pt]
        table[x=Params, y=Error]
        {data/S3DIS/Params/gen20.tex};
        \addplot[
            color=green]
        table[x=Params, y=Error]
        {data/S3DIS/Params/gen30.tex};
        \addplot[
            color=green,
            only marks,
            mark=halfcircle,
            mark size=2pt]
        table[x=Params, y=Error]
        {data/S3DIS/Params/gen30.tex};
        \addplot[
            color=blue]
        table[x=Params, y=Error]
        {data/S3DIS/Params/gen40.tex};
        \addplot[
            color=blue,
            only marks,
            mark=halfcircle,
            mark size=2pt]
        table[x=Params, y=Error]
        {data/S3DIS/Params/gen40.tex};
        \addplot[
            color=gray]
        table[x=Params, y=Error]
        {data/S3DIS/Params/gen50.tex};
        \addplot[
            color=gray,
            only marks,
            mark=halfcircle,
            mark size=2pt]
        table[x=Params, y=Error]
        {data/S3DIS/Params/gen50.tex};
        \addplot[
            color=orange]
        table[x=Params, y=Error]
        {data/S3DIS/Params/gen59.tex};
        \addplot[
            color=orange,
            only marks,
            mark=halfcircle,
            mark size=2pt]
        table[x=Params, y=Error]
        {data/S3DIS/Params/gen59.tex};
        \addplot[only marks, red, mark=*, mark size=2pt] coordinates {
        (5.00759700000000, 37.22)
        };
        \node at (axis cs:5.00759700000000, 33){\small RLA};
        \addplot[only marks, red, mark=*, mark size=2pt] coordinates {
        (2.4209, 42.7100)
        };
        \node at (axis cs:2.4209, 39){\small SAP-2};
        \addplot[only marks, red, mark=*, mark size=2pt] coordinates {
        (2.6954, 40.9775)
        };
        \node at (axis cs:2.6954, 36){\small SAP-1};
        \legend{{}{Gen 1},{},{}{Gen 10},{},{}{Gen 20},{},{}{Gen 30}, {}, {}{Gen 40}, {}, {}{Gen 50}, {}, {}{Gen 60}}
        \end{axis}
    \end{tikzpicture}
    \caption{The MIoU-Params POF developed over 60 generations by SSS3D on S3DIS with the complete search space.} 
    \label{SSS3D-ParamsS3DIS}
\end{figure}
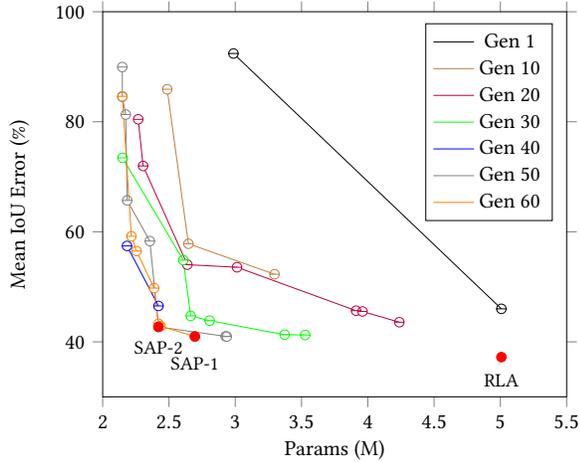

Figure~\ref{SSS3D-ParamsS3DIS} shows the evolution of the single-stage search POF over 60 generations of minimizing the mIoU error and the number of parameters.
Note that no fine-tuning is performed during search.
The original performance of the RandLA-Net supernet (RLA) does not appear in any POF since the individual included in the initial population used the weights resulting from progressive shrinking. 
PS weights result in higher error for RLA, as they are trained to increase the performance of every possible subnetwork.

Two models optimized for accuracy and parameters are highlighted in the figure, Sampling-Architecture-Parameters (SAP) -1 and -2.
They were found in the 60$^{th}$ and 50$^{th}$ generation respectively, and are detailed in Table~\ref{tableCompare}.
The trade-offs of SAP-1 and SAP-2 vs. RLA clearly indicate that RandLA-Net is over parameterized, and there is substantial room for improving efficiency.

Sampling-Architecture-FLOPs (SAF) networks presented in Table~\ref{tableCompare} were obtained during a single-stage search that minimized mIoU error and the number of FLOPs.
SAF-1 was discovered in generation 40, SAF-2 and SAF-3 in generation 60.
Their reduction in FLOPs show that point-based 3D semantic segmentation networks can be made more efficient.
It also helps tackling their biggest bottleneck which is memory accesses.


\subsection{Two-stage Search}

When SSS3D performs a two-stage search, first stage focuses on the sampling search space only and uses the FLOPs as the computational metric to optimize: the sampling hyperparameters have no effect on the number of parameters.
Since the architectural hyperparameters are not used, the progressive shrinking algorithm generally used for tuning the weights is not necessary for this stage, hence eliminating training altogether.

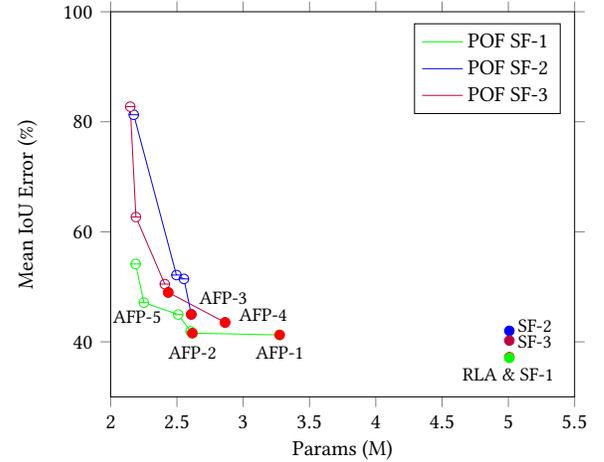
\begin{figure}[!t]
\centering
    \begin{tikzpicture}[scale=0.9]
        \begin{axis}[
            enlargelimits=false,
            xlabel={Params (M)},
            ylabel={Mean IoU Error (\%)},
            tick align=outside, legend pos=north east,
            xmin= 2, xmax=5.5,
            ymin= 30, ymax=100,
        ]
        \addplot[
            color=green]
        table[x=Params, y=Error]
        {data/S3DIS/Arch_1/gen15.tex};
        \addplot[
            color=green,
            only marks,
            mark=halfcircle,
            mark size=2pt]
        table[x=Params, y=Error]
        {data/S3DIS/Arch_1/gen15.tex};
        \addplot[
            color=blue]
        table[x=Params, y=Error]
        {data/S3DIS/Arch_2/gen10.tex};
        \addplot[
            color=blue,
            only marks,
            mark=halfcircle,
            mark size=2pt]
        table[x=Params, y=Error]
        {data/S3DIS/Arch_2/gen10.tex};
        \addplot[
            color=purple]
        table[x=Params, y=Error]
        {data/S3DIS/Arch_3/gen15.tex};
        \addplot[
            color=purple,
            only marks,
            mark=halfcircle,
            mark size=2pt]
        table[x=Params, y=Error]
        {data/S3DIS/Arch_3/gen15.tex};
        \addplot[only marks, red, mark=*, mark size=2pt] coordinates {
        (5.00759700000000, 37.22)
        };
        \addplot[only marks, green, mark=*, mark size=2pt] coordinates {
        (5.00759700000000, 37.09)
        };
        \node at (axis cs:5, 34){\small RLA \& SF-1};
        \addplot[only marks, red, mark=*, mark size=2pt] coordinates {
        (3.275, 41.26)
        };
        \node at (axis cs:3.275, 38){\small AFP-1};
        \addplot[only marks, red, mark=*, mark size=2pt] coordinates {
        (2.616, 41.58)
        };
        \node at (axis cs:2.616, 38){\small AFP-2};
        \addplot[only marks, blue, mark=*, mark size=2pt] coordinates {
        (5.00759700000000, 42)
        };
        \node at (axis cs:5.2, 43){\small SF-2};
        \addplot[only marks, red, mark=*, mark size=2pt] coordinates {
        (2.608, 44.99)
        };
        \node at (axis cs:2.85, 48){\small AFP-3};
        \addplot[only marks, purple, mark=*, mark size=2pt] coordinates {
        (5.00759700000000, 40.23)
        };
        \node at (axis cs:5.2, 40){\small SF-3};
        \addplot[only marks, red, mark=*, mark size=2pt] coordinates {
        (2.863, 43.52)
        };
        \node at (axis cs:3.15, 45){\small AFP-4};
        \addplot[only marks, red, mark=*, mark size=2pt] coordinates {
        (2.434, 48.98)
        };
        \node at (axis cs:2.2, 44.5){\small AFP-5};
        \legend{{}{POF SF-1},{},{}{POF SF-2},{},{}{POF SF-3}}
        \end{axis}
    \end{tikzpicture}
    \caption{The MIoU-Params POF of the selected subnetworks of the two-stage searches.} 
    \label{SSS3D-TwoStage}
\end{figure}

Two models highlighted in Figure~\ref{SSS3D-TwoStage} are Sampling-FLOPs (SF) -1 and -2, which are part of the POF of the 15$^{th}$; a third, SF-3, was discovered in the 20$^{th}$ generation.
SF-1 has a mIoU error of 37.09\%, which is lower than the original RLA’s 37.22\%, but with a FLOPs reduction of 1.9G, or 11.16\%.
A sampling strategy adjusted to a specific network architecture, as opposed to using the sampling values traditionally used in the field, can increase its performance while making it even more efficient.
Two-stage SSS3D can find these custom sampling strategies in less than 0.86 GPU days, compared to the 2.69 GPU days required for single-stage search.
SF-2 reduces the FLOPs of RLA by 10.26G, or 60.26\%, with an increase in mIoU error of 4.78\%.
SF-3, with its mIoU error of 40.23\%, records an increase of 3.01\% but with 44.99\% less FLOPs, a reduction of 7.66G FLOPs.

The second stage of the two-stage search takes subnetworks selected at the end of the sampling search (SF-1, -2, and -3) and starts architectural searches for each.
The sampling parameters of each search are fixed to the hyperparameters of the selected model (see Table~\ref{tableCompare}).
The architectural searches jointly optimize mIoU error and parameter count. 
This stage-wise approach reduces search time further, by 46\% on average, as the two smaller search spaces allow for a reduction in the population size and the number of generations.
The use of a different computational cost metric in the first and second stage implies that both of them are optimized, as well.
The performance of the SF networks at the beginning of the second stage, as well as the subnetworks that were found throughout the second stage, are shown in Figure~\ref{SSS3D-TwoStage}.
Architectural-FLOPs-Params (AFP) -1 and -2 were discovered on the 15$^{th}$ generation using SF-1. 
AFP-3 was obtained on the 10$^{th}$ generation using SF-2.
AFP-4 and AFP-5 were found in the 15$^{th}$ generation using SF-3.
The performance and characteristics of the subnetworks are presented in Table~\ref{tableCompare}.

Multi-objective search with three different metrics is possible, but at the cost of longer searches. 
It takes 3.24 GPU days to complete this search, while it takes 2.6 GPU days in total to complete the first stage once and the second stage for SF-1, -2 and -3.
Sampling-Architectural-Parameters-FLOPs (SAPF) networks, presented in Table~\ref{tableCompare}, were found by a single-stage search with mIoU error, the number of parameters, and the number of FLOPs, as search objectives, a population size of 15, and 80 generations.
When optimizing for the number of parameters and FLOPs simultaneously, the single-stage search needed 20 more generations to find networks competitive with two-stage search networks. 
Even so, they do not achieve similar FLOPs reduction. 
This is because the two-stage search has the ability to diversify the search or change its direction in between its two stages by choosing the sampling strategy for the second stage.

\subsection{Subnetwork Comparison}

\begin{table*}[t]
\centering
\caption{Characteristics of the networks resulting from the different searches. Comparisons with the supernet in parenthesis.}
\begin{adjustbox}{width=0.8\paperwidth, center}
\begin{tabular}
{|c|c|c|c|c|c|c|c|c|c|}
    \hline
    Name & Filter Ratio & Stride & K & Sub Sampling & Params (M) / (\%) & FLOPs (G) / (\%) & mIoU & FT mIoU / (\%) & GPU Days\\
    \hline
    \hline
    RLA & 1.0, 1.0, 1.0, 1.0, 1.0, 1.0, 1.0, 1.0, 1.0, 1.0, 1.0, 1.0, 1.0 & 1, 1, 1, 1, 1, 1 & 16, 16, 16, 16, 16 & 4, 4, 4, 4, 2 & 5.008 / (100) & 17.03 / (100) & \textbf{62.78} & 62.78 / (100) & -\\
    \hline
    SAP-1 & 0.8, 0.8, 0.8, 1.0, 0.4, 0.4, 0.4, 0.8, 0.8, 1.0, 1.0, 0.8, 0.8 & 1, 1, 1, 1, 3, 1 & 16, 20, 16, 16, 22 & 4, 4, 6, 4, 2 & 2.695 / (54) & 15.06 / (88) & 59.02 & \textbf{63.05} / (100.43) & 2.57\\
    \hline
    SAP-2 & 0.8, 0.8, 0.8, 0.6, 0.4, 0.4, 0.4, 0.8, 0.8, 1.0, 1.0, 0.8, 0.8 & 1, 1, 1, 1, 4, 3 & 16, 18, 16, 16, 20 & 4, 4, 8, 8, 2 & \textbf{2.421} / (48) & 12.00 / (70) & 57.29 & 61.51 / (98) & 2.14\\
    \hline
    SAF-1 & 0.8, 0.6, 1.0, 1.0, 1.0, 0.6, 1.0, 1.0, 1.0, 0.6, 1.0, 0.8, 1.0 & 1, 1, 1, 1, 1, 3 & 16, 16, 16, 16, 16 & 8, 4, 8, 8, 2 & 4.343 / (87) & 6.54 / (38) & 53.99 & 58.84 / (94) & 1.79\\
    \hline
    SAF-2 & 0.8, 0.6, 1.0, 1.0, 1.0, 0.6, 1.0, 0.6, 1.0, 0.8, 1.0, 0.8, 0.6 & 1, 1, 1, 1, 2, 3 & 16, 16, 16, 16, 16 & 6, 4, 4, 8, 2 & 4.250 / (85) & 9.95 / (58) & 59.19 & 61.98 / (99) & 2.69\\
    \hline
    SAF-3 & 0.8, 0.6, 1.0, 1.0, 1.0, 0.6, 1.0, 0.6, 1.0, 0.8, 1.0, 0.8, 1.0 & 1, 1, 1, 1, 2, 3 & 16, 16, 16, 18, 16 & 8, 4, 4, 8, 2 & 4.251 / (85) & 8.06 / (47) & 57.44 & 59.73 / (95) & 2.69\\
    \hline
    SAPF-1 & 0.8, 0.8, 0.8, 0.6, 0.4, 0.4, 0.6, 0.6, 1.0, 1.0, 1.0, 0.8, 1.0 & 1, 1, 1, 1, 1, 1 & 16, 16, 20, 16, 16 & 8, 2, 4, 8, 2 & 2.493 / (50) & 13.03 / (77) & 57.30 & 62.44 / (99) & 2.84\\
    \hline
    SAPF-2 & 0.8, 0.8, 0.6, 0.6, 0.4, 0.4, 0.6, 1.0, 0.6, 1.0, 1.0, 0.8, 1.0 & 1, 1, 1, 1, 1, 1 & 16, 16, 20, 16, 16 & 6, 4, 4, 8, 2 & 2.491 / (50) & 10.32 / (61) & 55.83 & 61.17 / (97) & 2.84\\
    \hline
    SAPF-3 & 0.8, 0.8, 0.8, 1.0, 0.4, 0.4, 0.6, 0.6, 0.6, 1.0, 1.0, 0.8, 1.0 & 1, 1, 1, 1, 1, 1 & 16, 16, 20, 16, 16 & 6, 2, 4, 8, 2 & 2.773 / (55) & 16.98 / (100) & 59.54 & 62.44 / (99) & 3.24\\
    \hline
    AFP-1 & 0.8, 0.8, 0.6, 0.8, 0.8, 0.4, 0.8, 0.8, 1.0, 0.8, 0.8, 0.8, 0.6 & 1, 1, 1, 1, 1, 1 & 16, 16, 20, 16, 16 & 4, 4, 6, 4, 4 & 3.275 / (65) & 14.30 / (84) & 58.74 & 62.62 / (100) & 1.21\\
    \hline
    AFP-2 & 0.8, 0.8, 1.0, 0.8, 0.4, 0.4, 0.4, 0.8, 1.0, 0.8, 0.8, 1.0, 0.6 & 1, 1, 1, 1, 1, 1 & 16, 16, 20, 16, 16 & 4, 4, 6, 4, 4 & 2.616 / (52) & 14.36 / (84) & 58.42 & 62.69 / (100) & 1.21\\
    \hline
    AFP-3 & 0.8, 0.8, 1.0, 0.8, 0.4, 0.4, 0.4, 0.8, 0.8, 1.0, 1.0, 0.6, 0.8 & 1, 1, 1, 1, 1, 3 & 16, 16, 16, 16, 16 & 8, 4, 8, 8, 2 & 2.608 / (52) & \textbf{6.41} / (38) & 55.01 & 59.28 / (94) & \textbf{1.04}\\
    \hline
    AFP-4 & 0.8, 0.6, 1.0, 1.0, 0.4, 0.4, 0.6, 0.8, 1.0, 1.0, 1.0, 0.8, 1.0 & 1, 1, 1, 1, 1, 1 & 18, 16, 16, 16, 16 & 8, 4, 4, 4, 2 & 2.863 / (57) & 8.86 / (52) & 56.48 & 60.40 / (96) & 1.46\\
    \hline
    AFP-5 & 0.8, 1.0, 0.8, 0.6, 0.4, 0.4, 0.4, 0.8, 0.8, 1.0, 1.0, 1.0, 1.0 & 1, 1, 1, 1, 1, 1 & 18, 16, 16, 16, 16 & 8, 4, 4, 4, 2 & 2.434 / (49) & 8.80 / (52) & 51.02 & 60.21 / (96) & 1.46\\
    \hline
\end{tabular}
\end{adjustbox}
\label{tableCompare}
\end{table*}

Table~\ref{tableCompare} compares the different networks obtained by the single- and two-stage searches, before and after fine-tuning.
On average, the networks have a 4.39\% difference between the mIoU obtained with the progressive shrinking weights and the mIoU after fine-tuning.
AFP-5 registered a rise of 9.19\% mIoU after fine-tuning.
This is probably due to the fact that during progressive shrinking architectures similar to the one of AFP-5 were not subsampled often, resulting in few contributions to weight updates. 
SAP-1, example predictions shown in Figure \ref{searchSpace}(b), is 0.27\% more accurate but requires only 53.81\% of RandLA-Net’s parameters and 88.43\% of its FLOPs.
The lowest parameter count was achieved by SAP-2 with 48.34\% of RandLA-Net’s parameters and the lowest FLOPs count was achieved by AFP-3 with only 37.64\% of RandLA-Net’s FLOPs.

Many subnetworks have filter ratios equal to 0.4 or 0.6 in the last encoder layer, the bridge layer, and the first decoder layer.
These are the blocks with the most initial features.
Since these subnetworks still achieve competitive mIoU scores after this heavy pruning, these layers are clearly overparameterized.
Aside from these layers, filter ratios of 0.6 or 0.8 appear frequently across the selected subnetworks, further indicating overparameterization.
RandLA-Net was designed using traditional power of two values for the number of filters.
This means that the number of filters in each consecutive layers in the encoder is increased by a power of two.
The results of the different searches of SSS3D shows that this method often leads to overparameterization and that choosing the right number of filters for a specific architecture can create more efficient networks while maintaining accuracy.

The subsampling ratio is the sampling hyperparameter that had the most impact on FLOPs and that varied the most.
It reduces the number of memory accesses done by the point-based subnetwork because fewer points are required for computation.
The $K$ value varied the most in the single-stage search that optimized for the number of parameters, suggesting that a bigger $K$ supports greater filter pruning.
Stride is constant in the first four encoder layers.
However, in the last encoder layer and in the bridge, stride varies a lot for both of the single-stage searches.
This implies that a lot of the feature points used in these layers could be omitted without having a huge effect on accuracy 
because a lot of information about neighboring points is encoded, reducing information loss.

Comparing the results of the single- and two-stage searches, the biggest takeaway is that accuracy after fine-tuning is very similar. 
SAP-1 achieves 63.05\% mIoU while AFP-1 and AFP-2 achieve 62.62\% and 62.69\% respectively.
The two-stage search is also better at finding efficient networks, as it optimizes both the number of parameters and the number of FLOPs.
For example, only one of the subnetworks found by the two-stage search has more than 3M parameters, while three of the single-stage subnetworks have more than 4M parameters, 
being optimized for FLOPs only.

As expected, the single-stage searches require the most time.
Search cost is limited by evaluation time and the size of the search space, which requires a large population and number of generations to produce good results.
Two-stage search reduces search time, but not the search space. 
The first stage of the two-stage search takes 0.86 GPU days; the second, up to 0.60 GPU days.
On average, two-stage search takes 54\% of the time required by a single-stage search optimizing for a single computational metric (FLOPs or parameters). 
In about half the time, the two-stage search optimizes both.
Similarly, a two-stage search takes only 43\% of the time of a single-stage search optimizing both metrics.

\section{Conclusion}

We presented Supernet Semantic Segmentation 3D, a fast multi-objective NAS framework that significantly cuts the search time for finding efficient and hardware-aware 3D semantic scene segmentation networks.
SSS3D uses NSGA-II to subsample candidates from a RandLA-Net supernet, and improve their performance over the course of generations through crossover and mutation.

SSS3D found two networks, SAP-1 and SF-1, that outperform the original RandLA-Net.
SAP-1 gains 0.27\% in mIoU with 53.81\% of RandLA-Net’s parameters and 88.43\% of its FLOPs.
SF-1 increases mIoU by 0.13\%, reducing FLOPs by 11.16\%.
For mIoU drops of 1.27\% and 3.5\%, SAP-2 uses only 48.34\% of RandLA-Net’s parameters, and AFP-3 needs only 37.64\% of the supernet’s FLOPs.
The two-stage searches used by SSS3D are faster and find better networks than single-stage searches: they minimize the number of parameters and the number of FLOPs simultaneously in just 54\% of the time required for a single-stage search of the complete design space.

\section{Acknowledgments}
This research was made possible by the support of: the Natural Sciences and Engineering Research Council of Canada (NSERC), though grant number CRDPJ 531142-18; and, Synopsys Inc.

\clearpage

\bibliographystyle{ACM-Reference-Format}
\bibliography{sample-tinyml}

\end{document}